\documentclass[letterpaper, 10 pt, conference]{IEEEtran}  
\usepackage{caption}
\usepackage{amsmath}
\IEEEoverridecommandlockouts
\usepackage{float}
\usepackage{graphicx} 
\usepackage{placeins} 
\usepackage{tabularx}
\usepackage{authblk}
\usepackage{hyperref}
\usepackage{cite}
\usepackage{amsmath}
\usepackage{amssymb}
\graphicspath{{./images/}}
\begin{document}
\title{\LARGE \bf
Volume-DROID: A Real-Time Implementation of Volumetric Mapping with DROID-SLAM
\vspace{-4mm}}

\author[1]{Peter Stratton\thanks{pstratt@umich.edu}}
\author[1]{Sandilya Sai Garimella\thanks{garimell@umich.edu}}
\author[1]{Ashwin Saxena \thanks{ashwinsa@umich.edu}}
\author[1]{Nibarkavi Amutha}
\author[1]{Emaad Gerami}
\affil[1]{Robotics Department, University of Michigan, Ann Arbor, USA}

\maketitle
\thispagestyle{empty}
\pagestyle{empty}

\begin{abstract}
This paper presents Volume-DROID, a novel approach for Simultaneous Localization and Mapping (SLAM) that integrates Volumetric Mapping and Differentiable Recurrent Optimization-Inspired Design (DROID). 
Volume-DROID takes camera images (monocular or stereo) or frames from a video as input and combines DROID-SLAM, point cloud registration, an off-the-shelf semantic segmentation network, and Convolutional Bayesian Kernel Inference (ConvBKI) to generate a 3D semantic map of the environment and provide accurate localization for the robot.
The key innovation of our method is the real-time fusion of DROID-SLAM and Convolutional Bayesian Kernel Inference (ConvBKI), achieved through the introduction of point cloud generation from RGB-Depth frames and optimized camera poses. 
This integration, engineered to enable efficient and timely processing, minimizes lag and ensures effective performance of the system.
Our approach facilitates functional real-time online semantic mapping with just camera images or stereo video input. 
Our paper offers an open-source Python implementation of the algorithm, available at \url{https://github.com/peterstratton/Volume-DROID}.
\end{abstract}

\medskip

\begin{IEEEkeywords} 
\textbf{\textit{Index terms} --- Simultaneous Localization and Mapping (SLAM), DROID-SLAM, Bayesian Kernel Inference (BKI) Mapping}.
\end{IEEEkeywords}
\smallskip

\section{Introduction}
Simultaneous Localization and Mapping (SLAM) is a fundamental mobile robotics problem in which a robot constructs a map of its environment and localizes itself within the map. 
Previous SLAM approaches have focused on using particle filters \cite{sim2005vision, grisetti2007fast}, extended Kalman filters \cite{huang2007convergence}, and graph-based optimization methods \cite{grisetti2010tutorial, thrun2006graph} to complete localization, but recently the focus has shifted to applying machine learning to graph-based SLAM techniques \cite{https://doi.org/10.48550/arxiv.2108.10869, zhu2022niceslam}. 

\vspace{0.5\baselineskip}
Our paper focuses on enhancing the mapping representation of DROID-SLAM (Differentiable Recurrent Optimization-Inspired Design SLAM) \cite{https://doi.org/10.48550/arxiv.2108.10869} by incorporating a volumetric mapping representation.
We propose the application of Convolutional Bayesian Kernel Inference Neural Networks (ConvBKI) \cite{wilson2022convolutional} to generate a 3D semantic map of the environment using camera poses optimized by DROID-SLAM.
By integrating a 3D semantic map, our approach improves the accuracy and completeness of environmental information for autonomous navigation.
We adapt DROID-SLAM for 3D semantic mapping, integrate it with a simulated robot platform, and evaluate its performance using simulation data.
The potential contributions of our work to mobile robotics and visual SLAM, particularly for 3D mapping and autonomous navigation, are significant.

\section{Related Work}
\subsection{Visual SLAM}
Recently, there have been significant improvements in SLAM systems, which are used to estimate the position and create a map of an environment. 
Visual Simultaneous Localization and Mapping (VSLAM) methods refer to using cameras for this purpose, and are preferred over Light Detection And Ranging (LiDAR) methods due to their lower weight, cost, and ability to provide a better representation of the environment. 
As a result, several VSLAM approaches have been developed using different types of cameras and have been tested in various conditions and on different datasets. 
This has led to a surge of interest among researchers and the development of multiple methods \cite{s22239297}. 
 
As DROID-SLAM receives images or frames from the video, it creates an image consistency graph where vertices correspond to images. 
If two images are at least three time steps apart and view the same scene, they are considered to be connected. This connection is referred to as an \textit{edge}.
Each new frame is then passed through a feature extraction neural network. 
The extracted dense features are used to compute a 4D correlation volume ($C_{ij}$) via dot product of the edges, where a correlation lookup operator is defined. 
The correlation lookup operator ($L_r$) is used to take a grid of image coordinates as inputs in order to output the correlation between the features at each coordinate. 

\begin{figure*}[ht]
  \centering
  \includegraphics[scale=0.3]{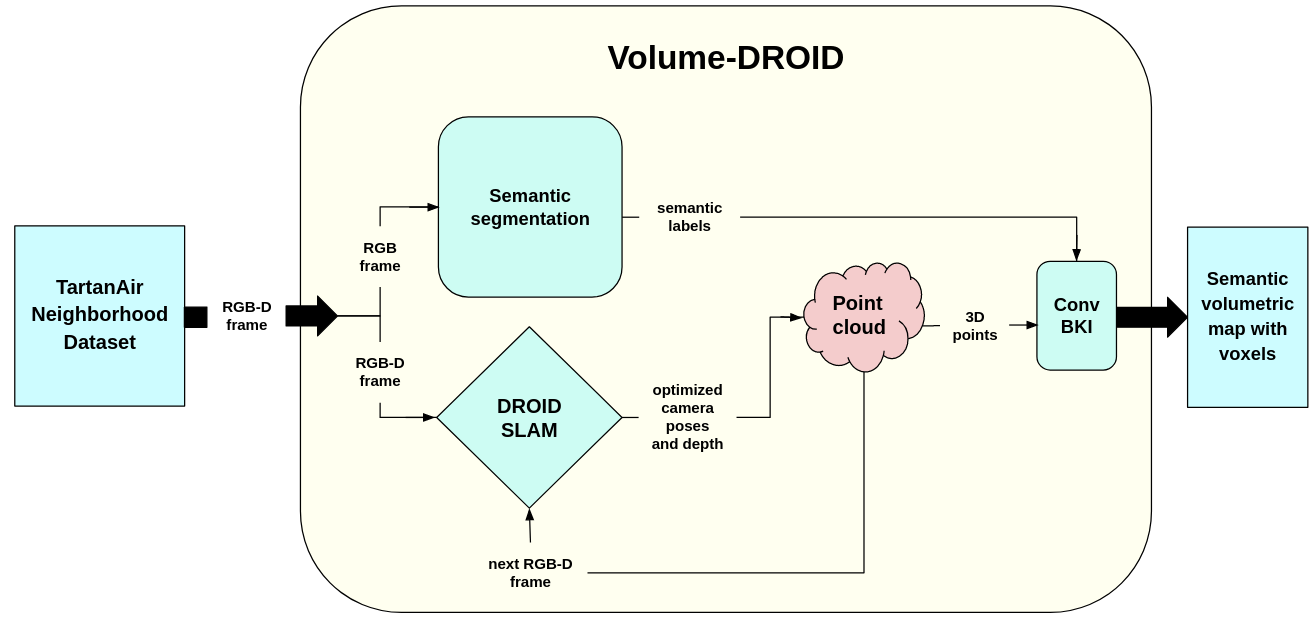}
  \caption{A visualization of the architecture of Volume-DROID.}
  \label{fig1}
\end{figure*}

\vspace{0.5\baselineskip}
From this point, the correspondence between each pair of frames in the image consistency graph is calculated and used to index the correlation volume as well as determine the optical flow between the two frames. 
This correspondence field ($p_{ij}$) consists of the camera model ($\prod_c$) mapping a set of 3D points onto the image and $\prod^{-1}_c$ is the inverse projection function mapping inverse depth map $d$ and coordinate grid $p_i$ to a 3D point cloud. 
The correlation features and flow features are each mapped through two Convolutional Neural Network (CNN) layers before being injected into a Gated Recurrent Update (GRU) \cite{cho-etal-2014-properties}. 

\vspace{0.5\baselineskip}
The "learning" of the DROID-SLAM occurs in the update operator and specifically in ConvGRU as the GRU has a gating mechanism that selectively updates and rejects information (features from feature vector) thus controlling information flow. 
The output of ConvGRU is revised flow field ($r_{ij}$) and its corresponding confidence weight ($w_{ij}$). 
The revised field information is concatenated with the context features from the poses and depths giving out corrected correspondence ($p^*_{ij}$). 
The corrected correspondence along with the confidence weight is passed in to Dense Bundle Adjustment layer (DBA) which maps the set of flow revisions into a set of pose and pixel-wise depth updates.
A sequence of these operations times the cost function $|E|$ results in optimized camera poses and optimized depth. It is to be noted that DROID-SLAM refers to depth as the inverse depth.

\vspace{0.5\baselineskip}

\subsection{Convolution BKI}
Convolutional Bayesian Kernel Inference (ConvBKI) is a differentiable 3D semantic mapping algorithm which combines reliability and trustworthiness of classical probabilistic mapping algorithms with the efficiency and optimizability of modern neural networks. 
ConvBKI layer explicitly performs Bayesian inference within a depthwise separable convolution layer. 
3D points are assigned semantic labels from off-the shelf semantic segmentation networks, and grouped into voxels by summing coinciding points. 
The constructed semantic volumes are convolved with a depthwise filter to perform a Bayesian update on a semantic 3D map in real-time.  
ConvBKI learns a distribution to geometrically associate points with voxels. 
A compound kernel enables ConvBKI to learn more expressive semantic-geometric distributions \cite{wilson2022convolutional}.

\subsection{Dilated Convolution for Semantic Segmentation}
To create a volumetric map, we need a semantic segmentation network. 
Older models for semantic segmentation are based on adaptations of convolutional networks that had originally been designed for image classification. 
However, dense prediction problems such as semantic segmentation are structurally different from image classification. 
Dilation convolutional networks are specifically designed for dense prediction. 
Dilated convolutions are used to systematically aggregate multi-scale contextual information without losing resolution. 
The architecture is based on the fact that dilated convolutions support exponential expansion of the receptive field without loss of resolution or coverage. 
Dilation convolutions networks have better accuracy than other state-of-the-art semantic segmentation models \cite{yu2016multiscale}.

\section{Methodology}
\subsection{Overall Process}
The process for Volume-DROID begins with utilizing the TartanAir neighborhood dataset, a virtual environment created in Unreal Engine, which eliminates the need to collect data for creating a custom environment.
Images or frames of the environment are then input to DROID-SLAM to generate optimized camera poses.
The optimized camera poses are used to generate a point cloud, with each pose stored as a 3D point. 
Subsequent images or frames are processed through DROID-SLAM to update the point cloud.
Finally, the collection of 3D poses is passed to ConvBKI, which converts them into a semantic volumetric map of the environment using voxels. 
Figure \ref{fig1} provides a visual representation of this process.

\begin{figure*}[ht]
  \centering
  \includegraphics[scale = 0.80]{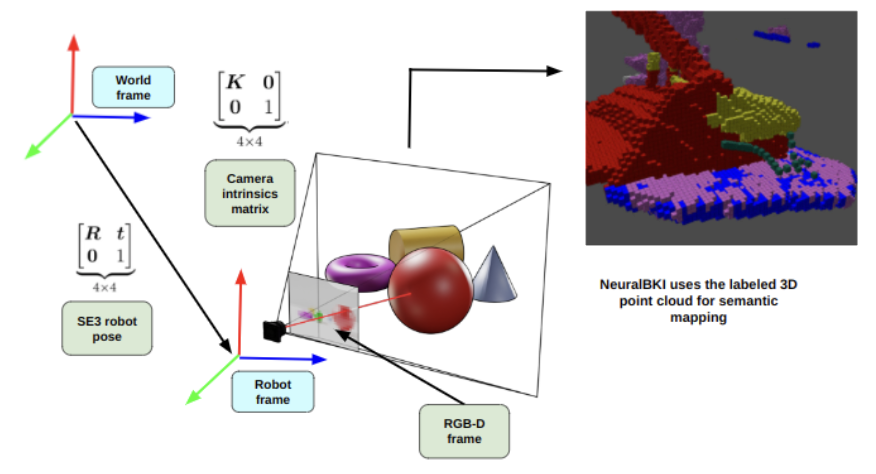}
  \caption{Visualization of point cloud projection and Convolutional BKI Process.}
  \label{fig2}
\end{figure*}

\subsection{From Pose and Depth to 3D Point Cloud}
To generate a 3D point cloud in the world frame, we combined the optimized SE(3) pose output from DROID-SLAM with the RGB-D frames obtained from TartanAir.
The process involves leveraging the SE(3) robot pose to establish the camera's position and orientation relative to the world frame.
The TartanAir RGB-Depth frames provide the necessary RGB and depth values, along with any additional information required for point cloud generation.
The camera intrinsic parameters, including the focal length and principal point, are also used to map the 2D image coordinates and convert them to 3D world coordinates.

\vspace{0.5\baselineskip}
The camera intrinsic parameter matrix, denoted as $K$, is represented by Equation \ref{eq:1}, where $f_x$ and $f_y$ correspond to the focal lengths in pixels along the $x$ and $y$ directions, respectively.
The camera frame centers are denoted by $c_x$ and $c_y$ along the $x$ and $y$ directions, respectively, and the skew coefficient, denoted as $S$, is assumed to be 0.
The matrix $K_h$ represents the homogeneous form of matrix $K$.

\begin{align}\label{eq:1}
    K &= \begin{bmatrix}
        f_x & S & c_x\\
        0 & f_y & c_y \\
        0 & 0 & 1
        \end{bmatrix},          
    K_{h} &= \begin{bmatrix}
            f_x & S & c_x & 0\\
            0 & f_y & c_y & 0\\
            0 & 0 & 1 & 0 \\
            0 & 0 & 0 & 1
            \end{bmatrix},
\end{align}

The primary matrix equation used in our point cloud projection is shown in \ref{eq:2}. 
The left-hand side (LHS) is the homogeneous representation of the point cloud coordinates, with $z$ representing the depth of each pixel from the $u-v$ depth map.
Equation \ref{eq:2} must be computed for each pixel in the $u$-$v$ depth map.

\begin{align}\label{eq:2}
         \begin{bmatrix}
           x \\
           y \\
           z \\
           1
         \end{bmatrix} &= z 
         \begin{bmatrix}
        R_{3\times3} & t_{3\times1}\\
        0_{1\times3} & 1
        \end{bmatrix}^{-1}
        \begin{bmatrix}
        K_{3\times3} & 0_{3\times1}\\
        0_{1\times3} & 1
        \end{bmatrix}^{-1}
        \begin{bmatrix}
           u \\
           v \\
           1 \\
           1/z
         \end{bmatrix}
\end{align}

Considering a resolution of 640 x 480 pixels for both depth and RGB frames, the computation in Equation (2) needs to be performed 307,200 times in total.
To reduce the computational burden, we derived the inverse of matrix $K_h$, denoted as $K_h^{-1}$, as illustrated in Equation (3).

\begin{multline}\label{eq:3}
    K_h^{-1} =
    \begin{bmatrix}
        K_{3\times3} & 0_{3\times1} \\
        0_{1\times3} & 1
    \end{bmatrix}^{-1} = \\
    \begin{bmatrix}
        1/f_x & -S/(f_xf_y) & (Sc_y - c_xf_y)/(f_xf_y) & 0 \\
        0 & 1/f_y & -c_y/f_y & 0 \\
        0 & 0 & 1 & 0 \\
        0 & 0 & 0 & 1
    \end{bmatrix}
\end{multline}

Instead of computing the 3D world coordinate for each $u$-$v$ pixel incrementally, we performed a batch computation on a version of Equation \ref{eq:2}, as shown in Equations \ref{eq:4} and \ref{eq:5}. 
The batch computation procedure begins with finding the matrix product of the SE(3) pose, the intrinsic values of the camera, and a batch matrix of $u$-$v$ pixels shown on the right-hand side (RHS) of Equation \ref{eq:4}.
Prior to the element-wise multiplication of the RHS of \ref{eq:4} with the corresponding $z_i$ matrix, we denote the world frame coordinates on the LHS of equation \ref{eq:4} with a $z-$ superscript.

\begin{multline}\label{eq:4}
    \begin{bmatrix}
    x_1 & x_2 & \dots & x_i\\
    y_1 & y_2 & \dots & y_i\\
    z_1 & z_2 & \dots & z_i\\
    1 & 1 & \dots &  1 
    \end{bmatrix}^{z-}_{4\times i=307200} \\
    =
    \begin{bmatrix}
        R_{3\times3} & t_{3\times1}\\
        0_{1\times3} & 1
    \end{bmatrix}^{-1}_{4\times4}
    \begin{bmatrix}
        K_{3\times3} & 0_{3\times1}\\
        0_{1\times3} & 1
    \end{bmatrix}^{-1}_{4\times4} \\
    \times
    \begin{bmatrix}
    u_1 & u_2 & \dots & u_i\\
    v_1 & v_2 & \dots & v_i\\
    1 & 1 & \dots & 1\\
    1/z_1 & 1/z_2 & \dots &  1/z_i
    \end{bmatrix}_{4\times i=307200}
\end{multline}

The final batch computation for obtaining world coordinates of each pixel from the $u$-$v$ depth map is shown in Equation \ref{eq:5}, where we apply element-wise multiplication (denoted by $\odot$) between the world coordinates prior to z-multiplication ($z-$ superscript matrix) and a row vector of corresponding $z_i$.

\begin{multline}\label{eq:5}
    \begin{bmatrix}
    x_1 & x_2 & \dots & x_i\\
    y_1 & y_2 & \dots & y_i\\
    z_1 & z_2 & \dots & z_i\\
    1 & 1 & \dots &  1
    \end{bmatrix}_{4\times i=307200} = \\
    \begin{bmatrix}
    z_1 & z_2 & \dots & x_i
    \end{bmatrix}_{1\times i=307200} \odot
    \begin{bmatrix}
    x_1 & x_2 & \dots & x_i\\
    y_1 & y_2 & \dots & y_i\\
    z_1 & z_2 & \dots & z_i\\
    1 & 1 & \dots &  1
    \end{bmatrix}^{z-}_{4\times i=307200}
\end{multline}

The matrix computation in Equation \ref{eq:5} is completed in real-time as the SE(3) pose and RGB-D frames enter this portion of the SLAM pipeline. 
Each point from the point cloud is then allocated a voxel based on ConvBKI.
Figure \ref{fig2} visually presents the point cloud projection and Convolutional BKI process, facilitating the reader's understanding of the operation sequence.

\section{Experiments And Results}
To evaluate our algorithm, we ran inference on the neighborhood section of the the TartanAir dataset \cite{tartanair2020iros}. 
The TartanAir dataset contains photo-realistic simulation environments in the presence of various light conditions, weather and moving objects.
DROID-SLAM is trained and evaluated on multiple environments within the TartanAir dataset. 
However, NeuralBKI was trained and evaluated on the KITTI dataset \cite{behley2019semantickitti}. 
to leverage preexisting models and avoid training from scratch, we evaluated Volume-DROID using the neighborhood subset of the TartanAir dataset. 
This subset was selected because it exhibits a higher degree of overlapping class labels with the KITTI dataset, compared to other subsets within TartanAir. 
By using the pretrained DROID-SLAM and NeuralBKI models, trained on the TartanAir and KITTI datasets respectively, we were able to effectively execute Volume-DROID.

\subsection{Implementation}
To demonstrate the proof of concept and evaluate the performance of Volume-DROID doign inference, our focus was on integrating trained DROID-SLAM, semantic segmentation, and NeuralBKI networks. 
However, for our experiments, we were only able to successfully integrate pretrained DROID-SLAM and NeuralBKI models. 
While we did not integrate a trained semantic segmentation network, it is important to note that our open-source Python implementation makes it straightforward to incorporate a trained semantic segmentation model.
Given the unavailability of a fully trained semantic segmentation network, we leveraged the semantic segmentation ground truth labels from the TartanAir dataset directly in ConvBKI. 
However, this approach posed a challenge due to the disparity between the semantic classes in TartanAir and the KITTI dataset. 
To mitigate this issue, we performed preprocessing on the TartanAir ground truth labels and mapped them to the corresponding KITTI class labels. However, this mapping process introduced some label noise since the neighborhood subset of TartanAir contains more semantic classes than KITTI. Consequently, TartanAir classes without a direct mapping to a specific KITTI class were assigned an arbitrary KITTI class label.
As a consequence of these factors, the semantic labeling exhibited some degree of noise, resulting in imperfect 2D semantic segmentation input for ConvBKI. 
Nevertheless, it is worth emphasizing that with a trained semantic segmentation network readily integrated into our open-source Python implementation, the overall performance of Volume-DROID in this aspect can be significantly improved.

\vspace{0.5\baselineskip}
We built Docker containers with publicly available Docker images to run and test Volume-DROID. 
This architecture makes Volume-DROID flexible to run on any operating system or environments which are able to run Docker. 
Our Docker image included pre-installed libraries necessary for running the components of Volume-DROID, namely Robot Operating System (ROS) Noetic and PyTorch. 
Since we ran everything on a server with no default Graphical User Interface (GUI), we had to use a noVNC (an open source Virtual Network Computing client) GUI display Docker container that could be accessed from any browser.
To visually inspect the 3D semantic map generated by Volume-DROID, we used ROS Visualization (RViz) within the GUI container. 
We used ROS to establish real-time communication between Volume-DROID and the ConvBKI functions, enabling the seamless exchange of optimized pose information.

\subsection{Qualitative Results}
\begin{figure}[h]
  \centering
  \includegraphics[scale = 0.55]{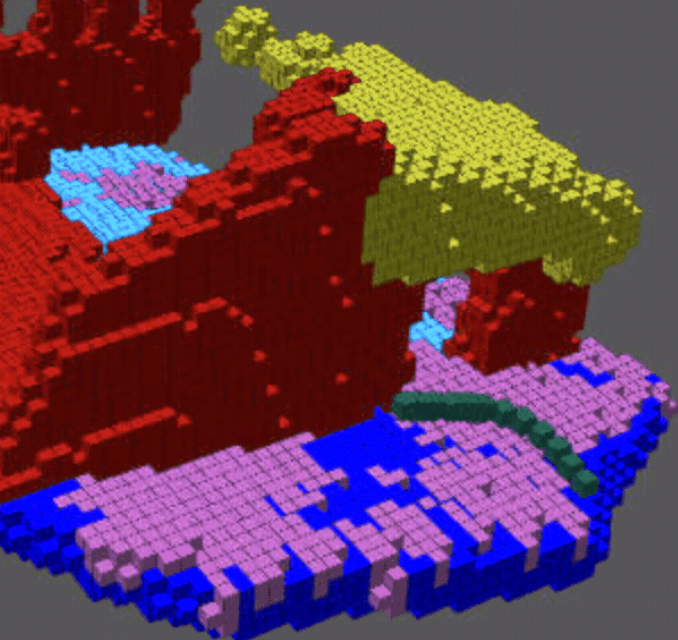}
  \caption{The end result of the Volume-DROID test.}
  \label{fig3}
\end{figure}

By using the optimized poses obtained from DROID-SLAM, we successfully visualized the voxels. 
However, we encountered difficulties in generating semantic segmentation results due to issues with the pre-trained model trained on the KITTI dataset. 
As shown in Figure \ref{fig3}, the RViz visualization of the Volume-DROID output reveals discernible outlines of shapes, such as the ground, building sides, and roofs. 
Nevertheless, the non-uniform 3D segmentation of the ground may be attributed to incorrect labels introduced during the TartanAir-to-KITTI class matching process. 
Furthermore, since ConvBKI was trained on KITTI data and not specifically evaluated on TartanAir data, its ability to generalize to the TartanAir dataset remains uncertain. 
Consequently, the patchy semantic labeling observed may be an artifact of ConvBKI's limited generalization to the TartanAir dataset.

\subsection{Quantitative Results}
DROID-SLAM measured three separate error scores during its tests. 
The first of these scores was the Absolute Trajectory Error (ATE), which can be described as a comparison between a robot's computed trajectory and its actual, traveled trajectory. 
The second score was the Relative Pose Error (RPE), which compared the reconstructed relative transformations between nearby poses to the actual relative transformations. 
The RPE returned two values due to the fact that it compares a pair of timestamps. The last of the three scores was the KITTI score, which provided the translational and rotational errors for all possible subspace lengths.
The values found in the tests can be found in Table 1. 
No semantic segmentation results were obtained, so no quantitative results are provided.

\begin{table}[h]
\begin{center}
\label{table:1}
\caption{DROID-SLAM Results}
\begin{tabular}{ | m{4cm} | m{3cm}| } 
  \hline
  Absolute Trajectory Error Score& 0.01755 \\ 
  \hline
  Relative Pose Error Score& 0.003769, 0.06087\\ 
  \hline
  KITTI Score& 0.01088, 0.002345\\ 
  \hline
\end{tabular}
\end{center}
\end{table}
\section{Discussion}
\subsection{Analysis of Results}

The analysis of the results indicates that DROID-SLAM performed well, demonstrating the effectiveness of the Volume-DROID algorithm. 
However, there are still opportunities for improvement within the system.
A limited time frame prevented us from fully training the algorithms.
Specifically, training DROID-SLAM alone would have taken a week using four RTX 3090s, which was not feasible within our project timeline. 
To overcome this constraint, we prioritized developing a proof of concept/minimum viable product that demonstrates a real-time workable pipeline. 
Therefore, the focus of this paper is on the construction of the pipeline and showcasing its functionality, rather than presenting fully trained models or quantitative results.
 Additionally, the time constraints hindered the implementation of a semantic segmentation network within the pipeline. 
 Integrating a trained semantic segmentation network would have potentially resulted in further improvements to the algorithm, leading to even lower error scores than those presented in Table 1.

\subsection{Future Work}
If more time and computational resource had been available, two key areas would have been the focus of attention: training a mapper for DROID-SLAM and retraining ConvBKI. 
Training a semantic mapper on DROID-SLAM's data could result in improved inference performance. 
Similarly, retraining ConvBKI would enhance the accuracy of its semantic map.
Following the completion of training tests for both DROID-SLAM and ConvBKI, the next step would be to evaluate Volume-DROID on a robot equipped with only a camera, as opposed to using TartanAir's purely simulation data. 
This real-world deployment would provide valuable insights into the algorithm's performance and robustness.
Additionally, integrating Volume-DROID with Hierarchical Representation \cite{ravichandran2022hierarchical} could be explored. 
This integration would enable a more abstract semantic representation of the environment, leading to improved planning capabilities and a better understanding of the environment for both humans and robots.
It is important to note that due to the limited training time available, these further improvements and integrations were not pursued in this project. However, they represent potential avenues for future research and development of Volume-DROID, through bridging the theory in this paper and our open-source implementation.

\section{Conclusion}
In this paper, we presented Volume-DROID, a novel approach for real-time simultaneous localization and mapping (SLAM) that integrates volumetric mapping with DROID-SLAM. 
Our approach combines DROID-SLAM, point cloud registration, an off-the-shelf semantic segmentation network, and Convolutional Bayesian Kernel Inference (ConvBKI) to generate a 3D semantic map of the environment and provide accurate localization for the robot.

We introduced the concept of real-time fusion of DROID-SLAM and ConvBKI through the generation of point clouds from RGB-Depth frames and optimized camera poses. 
This integration allows for efficient and timely processing, minimizing lag and ensuring effective performance of the system. 
By incorporating a 3D semantic map, our approach improves the accuracy and completeness of environmental information for autonomous navigation.

We provided a detailed overview of the methodology and the overall process of Volume-DROID, including the steps involved in generating a 3D point cloud from camera poses and RGB-D frames. 
We also discussed the significance of using dilated convolutions for semantic segmentation and the potential of ConvBKI for real-time 3D semantic mapping.

We believe that the contributions of our work to the field of mobile robotics and visual SLAM, particularly in the context of 3D mapping and autonomous navigation, are significant. 
The open-source Python implementation of the Volume-DROID algorithm further enhances its accessibility and encourages its adoption by the research community.

In future work, we plan to evaluate the performance of Volume-DROID using real-world data and conduct experiments to validate its effectiveness in various environments. 
We also aim to explore further optimizations and refinements to improve the efficiency and accuracy of the system.
Overall, we are optimistic about the potential of Volume-DROID to contribute to the advancement of SLAM techniques and enable more robust and reliable autonomous systems.
\addtolength{\textheight}{-12cm}   

\nocite{*}
\bibliographystyle{IEEEtran}
\bibliography{ref.bib}
\end{document}